# The Emerging Artificial Intelligence Protocol for Hierarchical Framework

Caesar Wu, PhD, Senior IEEE
Computer Science / SnT
University of Luxembourg
Luxembourg
https://orcid.org/0000-0002-2792-6466

Pascal Bouvry, PhD, IEEE
Computer Science / SnT
University of Luxembourg
Luxembourg
https://orcid.org/0000-0001-9338-2834

*Abstract*—Artificial intelligence (AI) advancements allow machines to achieve human-like intelligence. Problem-solving and decision-making are two mental abilities to measure human intelligence. Building a generalized computational model or representation for various inputs and outputs is essential to obtaining such human-like capabilities. Many scholars tried to articulate different models from different perspectives. However, there is a gap in establishing an overall AI-oriented hierarchical framework. Often, some outputs of AI processes could not be explained. This study proposes a novel model known as the emerged AI protocol that consists of seven distinct layers capable of providing an explainable solution for a given problem. In contrast to previous hierarchies, we argue that this unique model is conceptually evident, logically consistent, theoretically compelling, and practically adaptable. We aim to create a generalized model that can be implemented by various machine learning (ML) algorithms for problem-solving and decision-making.

## I. Introduction

Problem-solving and decision-making are considered to be two critical traits to measure people's intelligence because these two mental abilities reflect how well a person can cope with nature, society, and self in terms of challenges and responses. Russell [2] defined it as "an entity is considered to be intelligence… if it chooses (or decides) actions that are expected to achieve its objectives (or solving problems). In general, human intelligence is not an isolated talent but "a very general mental capability that, among other things, involves the ability to reason, plan, `solve problems, think abstractly, comprehend complex ideas, learn quickly and learn from experience."[1] That is to say, human-like intelligence is a kind of network or emerging phenomenon. The question is, how can we build a machine with human-like intelligence? Also, how can we explain or interpret the results produced by machines if we use the reverse logic of the ML mechanism (i.e. neural net)?

To answer these questions, it is critical to understand how our meat machine [3] organizes our mental representations of the complex world around us. Sternberg [4] proposed the theory of successful intelligence, which is the ability to understand complex social cultural contexts or social networks. He differentiates analytic and successful intelligence. Sternberg further qualified that the meaning of success is "attained through a balance of analytical, creative, and practical abilities" and highlighted that successful intelligence requires wisdom, which is the "power of judging rightly and following the soundest course of action based on knowledge, experience, understanding etc."

However, we often focus on the analytic side of intelligence, such as logical reasoning, increasing memory and computational power, pattern recognition, and optimizing algorithms in AI/ML practice. We hardly pay attention to codifying wisdom into an AI/ML program because "rightly" or "soundest" are subjective and hard to quantify. Nevertheless, it does not mean that many previous scholars have not had a trial to embed wisdom into a computational model. The Nobel laureate Simon [7] proposed the architecture (three levels of the framework) of complexity to extract common properties of different complex systems for general systems theory. (See Fig. 1A) Simon and Newell [6] [7] suggested that "there is a deeper beauty in the simplicity of underlying process that accounts for the external complexity."

Similarly, Minsky [8] introduced a human intelligence model, which he called the society of mind built from simple interactive elements known as mindless agents. Later, Minsky tried to mimic how the human mind works from a psychological perspective called an emotion machine [9]. He demonstrated a multilayer hierarchy of the human mind model. (See Fig.1B). Regarding psychology, Maslow[10] proposed a hierarchy of human needs (Refer to Fig. 1 C) from a human motivation and psychological perspective. By the same token, Ackoff [11] articulated a knowledge pyramid and loosely defined it as a thinking hierarchy built with data information, knowledge and wisdom (DIKW) from an educational perspective (See Fig. 1D).

Both Maslow's and Ackoff's pyramid is also divided into two segments, which are development and growth. However, these pyramid models drew many critics. Some [12] argued that the hierarchy is unsound because of logical error. Others [13] claimed Ackoff's hierarchy is irrelevant and dangerous because of the linearity of the continuum in terms of layers.

### A. Research problem, Motivation and Methods

By considering these critiques, this study will focus on the question of how we can articulate a generalizable representation model for different problem-solving strategies. Our goal is to develop a universal model that contains not only knowledge but

This research was funded in whole, or part, by the Luxembourg National Research Fund (FNR) grant reference.

ICOIN 2023 The 37th International Conference On Information Networking (ICOIN 2023)

also intelligence, wisdom, and belief for problem-solving and decision-making strategies that can be explainable and understandable by humans. The motivation is to continue the research journey that Simon, Newell, Minsky and Marr [15] had initiated. Throughout this paper, we employ the isomorphism method to build our model. We use exploratory and descriptive methods to formulate the AI protocol in detail and then test it.

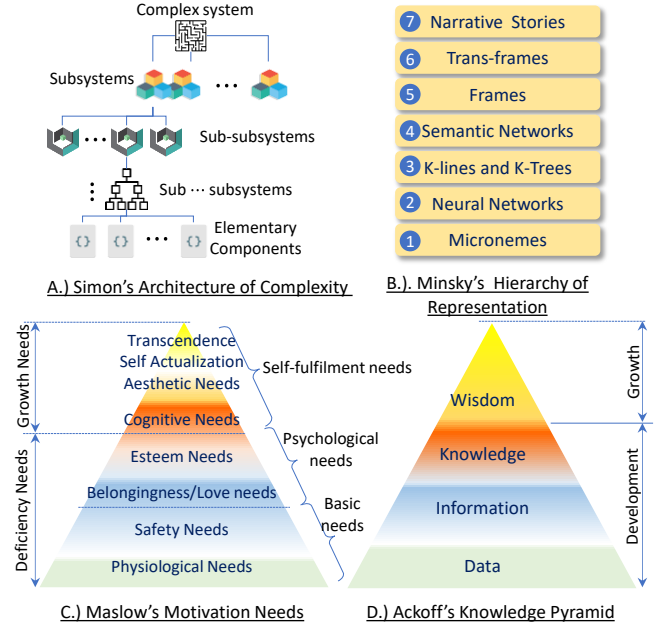

Fig. 1. Various Hierarchical Architectures

## B. Contributions

By creating the emerged AI protocol, the research made the following contributions:

- To the best of our knowledge, it is the first time established a generalizable AI protocol.
- This study clarifies many essential AI concepts that we often use them interchangeably and ambiguously.
- The paper formulates a hierarchical framework that consists of seven distinct layers.
- We use some generalizable mathematical equations to represent the characteristics of each layer. Therefore, it proposes a new representation model for AI/ML algorithms.
- We use one of the text mining examples to illustrate the implication of the protocol, which can be applied to different domains of AI.

## C. Scope of The Research

The rest of the paper consists of five sections: Section 2 gives a brief literature review of various hierarchical representation models. Section 3 presents the emerged AI protocol. Section 4 is to apply the AI protocol for a text mining analysis. Section 5 discusses the emerged AI protocol's meaning and rationality by comparing it to existing representation models. The final section provides our conclusions and future direction.

## II. LITERATURE REVIEW

The idea of hierarchy has some Platonic heritages, namely, the divided line. It is the central metaphor that Plato brought from mathematics in his landmark work – the Republic. Plato used the divided line to describe the relationship between imprecise statements from our sense experiences and the exact statements in the eternal and unchanging reality. From Plato's origin, the hierarchical models can be classified into four types: complex system hierarchy, Agent-based hierarchy, Pyramid-based hierarchy, and stack-based protocol or hierarchy.

## A. Complex System Hierarchy

Simon [5] intends to find a general and simple pattern behind complex systems. Simon's architecture has four attributes: 1.) The model forms multiple layers. 2.) It evolves. 3.) It has dynamic prosperities, 4.) It can be almost decomposed because most complex structures are enormously redundant. Simon's colleague Newell [14] also presents a function-based model that illustrates a general intelligent agent who can systematically represent knowledge by leveraging the six layers of a computer system: device, circuit, logic circuit sublevel, register-transfer sublevel, program (symbol), and configuration. Newell's hierarchical model divides a computer system into two levels of abstraction: knowledge and symbol because knowledge is the abstract of "states of mind." It allows us to develop different AI technologies separately. Newell argued that knowledge is a radical approximation and must be represented through the symbolic level. Together with Newell and Shaw, Simon [7] laid out the foundational theory for AI development.

## B. Agent-Based Hierarchy

Along with a similar line of reasoning, Minsky [8] proposed a hierarchical model for the human mind (or intelligence) from an agent-based perspective. Minsky emphasized the emerged properties. It implies that the higher level entities can be emerged from lower-level agents by reducing the uncertainty. Later, Minsky developed a hierarchical model to illustrate how the human brain organizes knowledge at multiple levels from a learning perspective. He argued that the hierarchy architecture is the simplest arrangement for some large-scale representation models. Likewise, Marr [15] developed a mental representation model for visual information processing. It consists of three layers. The first layer is characterized as what is being computed for a goal. The second layer is an algorithmic layer, which is a way to achieve the goal. There are many possible means to reach the end. The third layer is to be realized physically. It is how to execute or implement algorithmic processes.

## C. Pyramid and Analytic Hierarchical Process (AHP)

Similarly, Maslow [10] hypothesized a pyramid hierarchy based on human motivation needs. He classified them into deficiency and growth needs. Maslow argued that if deficiency needs are absent, humans will generate unpleasant feelings. However, many people have not been convinced because 1.) Many needs do not follow a hierarchy. 2.) The hypothesis is hard to be falsified.

Like Maslow, Ackoff [12] presented a similar pyramid hierarchy from an educational perspective. Ackoff aimed to teach students how to systematically think or infer what they do not know based on a value proposition. Ackoff divided the layers of his pyramid into two classes: development and growth, in which the bottom three are donated to development, and wisdom belongs to growth. Ackoff asserted that development does not require value. Only wisdom deals with values or effectiveness. However, Jennex et al.[16] disagreed with Ackoff because they claimed that the number of combinations at the information layer should be greater than at the data layer. Likewise, knowledge is more than information. However, Ackoff's pyramid is about statistical abstraction rather than numerical combinations. Over the years, many scholars and practitioners have developed various models for decision-making and problem-solving, such as hierarchical learning [17], deep hierarchical learning [18], decision hierarchy with analytic hierarchical process (AHP) [19], analytic network process (ANP) [20], hierarchical abstraction [21], hierarchical complexity [22] Although some terms could be slightly different, the essential meaning of a model remains the same, such as higher-order representation [23], higher-order theories of consciousness [24], high-order theory of mind [25] or protocol stack.

### D. Stack-based Protocol and Protocol Analysis (PA)

During the 1990s, a stack protocol was developed for Internet communication, which is "a set of rules that define how systems interact"[26]. The internet protocol is a language of computer networking. One software system is the protocol analysis (PA) system (or PAS-I). It was originated by Waterman and Newell [27] for AI in the early 1970s. Although some scholars [28] preferred PA, others [29] argued that each method has pros and cons. They suggested that the best approach is to adopt a wide range of complementary methods. It inspires us to propose the emerging AI protocol.

### III. DEVELOPING EMERGED AI PROTOCOL

Our research problem is improving Ackoff's knowledge pyramid for computational AI. This question leads us to formulate an AI protocol to implement various algorithms based on intelligence, wisdom, and beliefs.

### A. Multilayers of Protocol and Isomorphism

The AI protocol emerged from two intellectual sources: the knowledge pyramid and Open System Interconnection (OSI) reference model. The new stack-based hierarchy consists of seven layers: bit, data, information, knowledge, intelligence, wisdom, and belief (See Fig.2). Compared with Ackoff's pyramid, the model embeds three new layers (bit, intelligence, and belief) into the hierarchy. Furthermore, instead of pyramid architecture, we drew an isomorphism from the OSI model to create the AI stacks. The following sections give more details:

### B. Defining Seven Layers of the Emerged AI Protocol

*1) Bit Layer and Sensory Signal*

A bit is the universal currency of both data and information. Lexically, a bit means binary digit. Tukey [30] first coined the term "bit." One bit means some possibility between two equally likely options. The core of a bit is not about its meaning but the ability to differentiate among various bits. The basic unit of the Turing machine is a bit. Binary computations are the fundamental expression of Turing machines. Shannon [31] used the entropy or H in Eq.(1) to represent the member of the message we can receive through a communication channel.

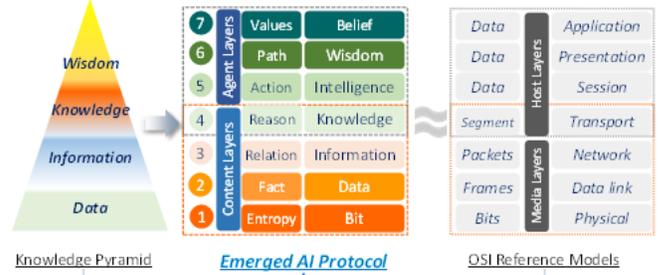

Fig. 2. Seven Layers of the Emerged AI Protocol

$$H = log_2 M = \frac{lnM}{ln2} \quad (1)$$

Where M represents the amount of message, if we only receive a one-bit signal or M=1, the entropy is equal to 0 bit because the message is 100% certain. Eq.(1) presents a generalized theorem of the bit layer. However, If we receive a string of bits, it means we can extract patterns of bits or data.

*2) Data Layer*

The concept of data is quite vague. We often use both data and information interchangeably. Ralph Hartley once said: "information is a very elastic term"[32]. So, it is the data. Data is "facts and statistics collected together for reference or analysis." Shannon uses the average entropy with different probabilities to represent the received data shown in Eq. 2

$$H(x) = \sum_x p(x)log_2(1/p(x)) \quad (2)$$

Where "X" is a variable, H(X) is entropy, and p(x) is the probability of a particular event "x" occurring. When we leverage the logical relation of data, the information emerges.

*3) Information layer*

Information can be defined as "data in a form." It is the "facts provided or learned about something or someone." If the data is unorganized facts, then the information is structured data. Janich argued, "Information is a natural object." [33] It is also something for us to communicate with its content. There is a conditional entropy H(T|R) that fails to communicate between transmitter: T and receiver: R. Eq. (3) represents the relationship between input T and output R.

$$H(T|R) = H(T \cup R) - H(T) \quad (3)$$

Shannon's information theory only focuses on syntactic structure rather than semantic meanings. The question of how to communicate information contents is the issue of knowledge.

*4) Knowledge Layer*

Plato's definition of knowledge is "justified true belief." Justification means supporting evidence learned from the world through testimony. Empirically, knowledge includes "know-how" and "knowledge-that." Eq. 4 represents the justification of true belief [34]

$$J(B_e|t_1, t_2, \cdots, t_k) = \frac{p(B_e \cap t_1 \cap t_2 \cap \cdots, t_k)}{[p(t_1)p(t_1|t_2, \cdots, t_k)]} \quad (4)$$

Where J is justification and, $B_e$ is a personal belief and, t is the testimony, p is the probability. We can think knowledge has a duality of both theoretical and practical properties. Immanuel Kant approximately said, "theory without practice is empty, and practice without theory is blind. "This statement leads to the question of how we can sensibly apply our learnt knowledge to the external world. It leads to the intelligence layer.

*5) Intelligence layer*

Based on Sternberg, intelligence means "the ability to carry on abstract thinking, the ability to learn to adjust to the environment,…the capacity for knowledge, the amount of knowledge possessed, and the capacity to learn or to profit from the experience."[35] Intelligence is the ability to transfer knowledge into executable actions and forces. Wissner-Gross and Freer [45] define intelligence in Eq.5

$$F(X_0) = T\nabla_X S(X)|_{X_0} \quad (5)$$

Where $F(X_0)$ is the intelligence force with the present macrostate $X_0$. "T" is the reservoir temperature or strength, and S(X) is the entropy associated with macrostate X, which are possible accessible futures. Notice that S(X) is equivalent to H(X) in Eq. (2). However, intelligence may slip into three paradoxes: 1.) The trap paradox implies that intelligent people often become arrogant [36] 2.) The strange preference paradox is that intelligent people often want to contest novelty ideas because they have more brain capacity.[37] 3.)The choice paradox suggests that intelligent people have more choices than less intelligent people because they think they deserve to maximize the value of their decisions.[38] The issue of how to avoid intelligent traps gives rise to wisdom.

*6) Wisdom Layer*

The connotation of wisdom often implies various spiritual and philosophic traits. It means having rich experiences, full knowledge, and sound judgments. Compared with intelligent judgment, a wiser decision may reveal a long-term consequences. If intelligence seeks more (interests) for less (efforts), then wisdom seeks less (a few ends) for more (more prosperous meaningful life). When we use wisdom to predict the future, we use many models that allow us to cancel out opposite errors. This is the essence of "The wisdom of the crowd."[39]. Page [40] proposed Eq.6 for the wisdom of the crowds.'

$$(C - P)^2 = \frac{1}{N}(\sum_{i=1}^{N}(x_i - P)^2 - \sum_{i=1}^{N}(x_i - C)^2) \quad (6)$$

Where "C" is the crowd's average predicted error, "P" is the actual prediction value and "$x_i$" is the individual predicted value. "N" is the number of individuals. The value $(C-P)^2$ means the wisdom of crowds. Overall, Sternberg summarized wisdom has three different senses: 1.) Sophia, which is to search for the truth in the contemplative life. 2.) Phronesis, which is knowing how to balance conflicting aims and principles. 3.) Episteme, which is to understand things from a scientific perspective. Phronesis is practical wisdom. It is how to dialectically balance different competing values and principles in practice. It depends on belief.

*7) Belief layer*

Simply put, belief is to accept something true without justification. Nilsson [41] depicts that "beliefs constitute one of the ways we describe the world we live in.". "Reality is out there…independent of what we believe about it, but we never experience that reality directly." Fortunately, rational believers will analyze and modify their beliefs before they will trust a belief sufficiently to act on it by considering their own experiences, reasoning, and the opinions and others' criticisms. Dempster-Shafer Theory (DST) [42][43] or belief functions mathematically present Nilsson's description shown in Eq.7 and Eq.8.

$$Bel(A) = \sum_{B \subseteq A} m(B), \quad \sum_{A \subseteq \Theta} m(A) = 1 \quad (7)$$

$$Pl(A) = \sum_{B \cap A \neq \phi} m(B) = 1 - Bel(\bar{A}) \quad (8)$$

$$\forall A, B \subseteq \Theta, \quad \Theta = \{b_1, b_2, \cdots, b_n\}$$

Where *"A"* is a hypothesis or proposition, or event, *"B"* includes all the focal elements that are a subset of Θ within the hypothesis *"A."* "*Θ*" represents a frame of discernment, which is a finite set of mutually exclusive elements in a domain. {$b_1$, $b_2$, …, $b_i$ } are the focal elements on which one has non-zero belief masses (e.g., $b_1$, or {$b_1$,$b_2$} is a focal element if $0<m(b_1)<1$ or $0<m(\{b_1,b_2\})<1$). $m(B)$ is a mass function, also known as a basic belief assignment. The advantage of DST equations is that they can effectively mitigate all positive or negative beliefs, while Eq.6 can only balance a mixture of positive and negative values.

IV. EXAMINE THE AI PROTOCOL THROUGH TEXT MINING

Text mining analysis is one of the most compelling examples to test the AI protocol because comprehensive text mining involves sensory inputs, symbolic manipulation, syntactic and semantic analysis, logic inference, contextual inquiry, meta-analysis and transcendental meditation. These analytic activities correspond approximately to the seven layers of the AI protocol.

*A. Example of Analyzing Random selected 36 papers*

To test the protocol, we randomly selected 36 recently published papers as the input dataset on the topic of trustworthy AI. The computational goal is to discover some of the most correlated papers for further analytic reading.

We can skip through the bit layer because the input data (36 files) is in digital format. If the dataset is collected from sensory inputs, we might have to extract data patterns from a bit string. In this case, we examine each paper's most frequent words, except for some stop words and numeric numbers (See Fig.3). The horizontal axis is the number of word accounts, and the vertical axis is the top 10 frequent words. At first glance, we can see that some paper is much longer than others. We argue that Fig.3 only shows average entropy. It does not deliver too many messages to us. The next step is how to do a cross-examination, which demonstrates each word across 36 selected papers versus within each paper. It allows us to compare strong deviations between individual papers with other papers. As we can see in Fig.4, words that are close to the diagonal dash line have similar frequency across all papers. Words standing above the line are common across all papers, and words below the line are common in a particular paper but not across other papers. This comparison test illustrates some conditional entropy based on Shanon's information theory (See Eq.3). Moreover, we quantify how similar and different these sets of words are by a correlation test of these papers.

We can consider this correlation test as "justified true belief" for us to select a few seminal papers to do the analytic reading.

This test corresponds to the knowledge layer. The top 5 most correlated papers are shown in Table I.

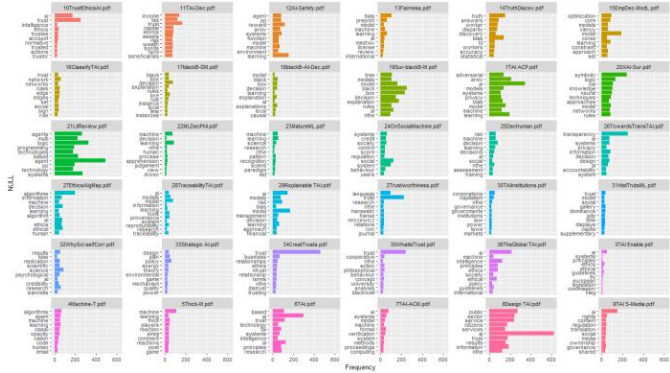

Fig. 3. Randomly Selected 36 Papers

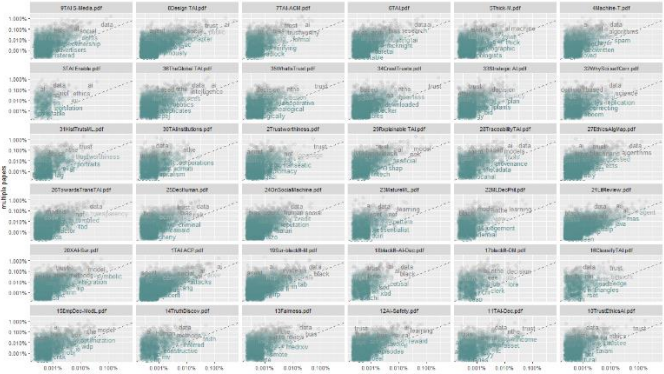

Fig. 4. Words Across All Papers

TABLE I. RESULT OF TOP 5 CORRELATION PAPER FROM 36

| Paper Title | Correlation | p-value |
|---|---|---|
| Trustworthy artificial intelligence | 0.754 | 0 |
| Trustworthy AI: A Computational Perspective | 0.749 | 0 |
| Guidelines for Designing Trustworthy AI… | 0.709 | 0 |
| Financial Risk Management and Explainable TAI | 0.704 | 0 |
| Trustworthy AI | 0.653 | 0 |

### B. Aggregating Result of 324 Papers

The intelligent layer is much more complex because we have to interact with the external world. In this case, we have read through each paper listed in Table-I to extract semantic meanings because a machine would not understand the semantic meaning or contents of each paper. However, we can gather more papers and divide them into a number of clusters and select the most correlated papers from each cluster and then re-iterate the process of the correlation test. This iteration is similar to the process of the wisdom of crowds. Another common term is meta-analysis. We implemented the experiment by collecting 324 papers and dividing them into 9 clusters. Table II shows the top 5 most correlated papers from 324 selected papers.

This simple experiment reflects the wisdom of crowds or demonstrates the meta-analysis. The issue of how "randomly" selecting papers depends on our beliefs, which is our experience. We use previous experience to pick up some keywords and then update these keywords when we search along. It is an iterative process driven from both top-down and bottom-up, which Eq.7 and 8 illustrate as a generalized principle.

TABLE II. AGGREGATED RESULT OF TOP 5 FROM 324 PAPERS

| Paper Title | Correlation | p-value |
|---|---|---|
| ITI's Global AI Policy Recommendations | 0.929 | 0 |
| Preparing for the future of AI | 0.925 | 0 |
| AI Now 2017 Report | 0.923 | 0 |
| Dutch AI Manifesto | 0.900 | 0 |
| Ethical Social and Political Challenges of AI in Health | 0.891 | 0 |

## V. DISCUSSION AND COMPARISON

This study proposes the emerged AI protocol, primarily derived from Simon's architecture complexity, Minsky's hierarchical representation and mental process hierarchy, David Marr's three levels of Vision, Ackoff's knowledge pyramid, and Maslow's motivation needs. We use the isomorphism method to create a model that has similarities to the OSI reference protocol. Compared with other models, we clarify the basic terms of the emerged AI protocol at different sophisticated levels and generalize some inputs and outputs of the ML process.

### A. Meaning of the Emerged AI Protocol

In our everyday life, we do not differentiate the terms "data mining", "information processing", "knowledge retrieving", and "computational intelligence." We use these terms interchangeably. It is easy to be confused when applying different AI/ML processes. By formulating the AI protocol, we assign various degrees of certainties to various algorithmic processes. It allows us to understand how much effort we should invest in what kind of problem and what kind of result we expect. Most importantly, the new AI protocol can offer a big picture of the problem-solving and decision-making strategy.

### B. Making Difference

By proposing this novel AI protocol, we made fundamental differences from both theoretical and practical perspectives:

- The protocol establishes a generalized framework for building various AI processes with different degrees of certainty.
- Practically, we can leverage this new protocol to create standard procedures to solve different levels of problems.
- If we consider problem-solving or decision-making as a metaphor for manufacturing, these layers become different input materials for the manufacturing operation.
- Instead of a purely philosophical argument, we articulated the mathematical equations for each layer of the protocol.

### C. Logic and Rationality

The logical implication is that our reason (it) or rationality not only emerges from a physical bit but is also derived from a virtual being (ideas). Without our ideas, there will be no "it." That is why we borrow John wheeler's slogan: "it from bit" [44], and create our own slogan: "it from being". Although the protocol has seven layers, we can classify them into three levels of stacks to deal with three kinds of problem landscapes: Mt Fuji, Rugged, and Dancing floor landscape. We argue that the first four layers correspond to the Mt Fuji type of problem because these layers can handle a relatively stable situation. Intelligence and wisdom layers deal with rugged landscape problems because a problem solution is interactive. We must trust our intuition or beliefs with a dancing floor problem.

## VI. CONCLUSION AND FUTURE DIRECTION

We have presented a hierarchical framework known as the emerged AI protocol that consists of seven layers. This new representation model is derived from subjective values and objective strings. The representation at each layer can be seen as pattern abstraction. The new protocol can fit with the existing data mining, information processing, and knowledge-extracting concepts. This novel protocol should advance the new way of the AI representation model. This study is one part of the theoretical framework for our research project. We do not only apply it to text mining analysis but also adopt it for real-world problems. This novel AI protocol could help us find elegant solutions for real-world challenges beyond our imagination.

ACKNOWLEDGEMENT

Fonds National de la Recherche (FNR) is the primary funder for this research

REFERENCES

[1] Gilovich, Thomas, Dale Griffin, and Daniel Kahneman, eds. Heuristics and biases: The psychology of intuitive judgment. Cambridge university press, 2002.
[2] Russell, Stuart, Human Compatible Artificial Intelligence and the problem of control, Viking, 2019.
[3] Clark, Andy. Mindware: An introduction to the philosophy of cognitive science. Oxford University Press, 2000.
[4] Sternberg, Robert J. Wisdom, intelligence, and creativity synthesized. Cambridge University Press, 2003.
[5] Simon, Herbert A., The architecture of complexity, Facets of systems science. Springer, Boston, MA, 1991, p. 457-476
[6] Newell, Allen, and Herbert Alexander Simon. Human problem-solving. Vol. 104, no. 9. Englewood Cliffs, NJ: Prentice-hall, 1972.
[7] Newell, Allen, John Calman Shaw, and Herbert A. Simon. "Elements of a theory of human problem-solving." Psychological review 65, no. 3 (1958): 151.
[8] Minsky, Marvin. Society of mind. Simon and Schuster, 1988.
[9] Minsky, Marvin. The emotion machine: Commonsense thinking, artificial intelligence, and the future of the human mind. Simon and Schuster, 2007
[10] Maslow, Abraham H. "The farther reaches of human nature." (1971)
[11] Ackoff, Russell L., From data to wisdom, Journal of applied systems analysis 16.1, 1989: p.3-9
[12] Frické, Martin., The knowledge pyramid: a critique of the DIKW hierarchy., Journal of information science 35.2, 2009, p.131-142
[13] Snowden, David J., and Mary E. Boone., A leader's framework for decision-making. Harvard business review 85.11, 2007, p. 68
[14] Newell, Allen. "The knowledge level." Artificial intelligence 18, no. 1 (1982): 87-127.
[15] Marr, David, "Vision A computational Investigation into the Human representation and processing of visual information, The MIT Press, 2010
[16] Jennex, Murray E., and Summer E. Bartczak. "A revised knowledge pyramid." International Journal of Knowledge Management (IJKM) 9, no. 3 (2013): 19-30.
[17] Bouvrie, Jacob V. "Hierarchical learning: Theory with applications in speech and vision." PhD diss., Massachusetts Institute of Technology, 2009.
[18] Anselmi, Fabio, Joel Z. Leibo, Lorenzo Rosasco, Jim Mutch, Andrea Tacchetti, and Tomaso Poggio. "Magic materials: a theory of deep hierarchical architectures for learning sensory representations." CBCL paper (2013): 16.
[19] Saaty, Thomas L., and Luis G. Vargas. Models, methods, concepts & applications of the analytic hierarchy process. Vol. 175. Springer Science & Business Media, 2012. p.23-60
[20] Saaty, Thomas L., and Luis G. Vargas. Decision making with the analytic network process. Vol. 282. Springer Science+ Business Media, LLC, 2006
[21] Kulkarni, Tejas D., et al. "Hierarchical deep reinforcement learning: Integrating temporal abstraction and intrinsic motivation." Advances in neural information processing systems. 2016
[22] Peinke, Joachim, et al. Encounter with chaos: self-organized hierarchical complexity in semiconductor experiments. Springer Science & Business Media, 2012
[23] Lloyd, John W. "Knowledge representation, computation, and learning in higher-order logic." Available at http:/csl. anu. edu. au/~ jwl (2001)
[24] Gennaro, Rocco J., ed. Higher-order theories of consciousness: An anthology. Vol. 56. John Benjamins Publishing, 2004
[25] De Weerd, Harmen, Rineke Verbrugge, and Bart Verheij. "Higher-order theory of mind in the tacit communication game." Biologically Inspired Cognitive Architectures 11 (2015): 10-21.
[26] Loshin, Pete. TCP/IP clearly explained. Elsevier, 2003. 13-14
[27] Waterman, Donald A., and Allen Newell. "Protocol analysis as a task for artificial intelligence." Artificial Intelligence 2, no. 3-4 (1971): 285-318.
[28] Ericsson, K. Anders, and Herbert A. Simon. Protocol analysis: Verbal reports as data. The MIT Press, 1984
[29] Burge, Janet E. "Knowledge elicitation tool classification." Artificial Intelligence Research Group, Worcester Polytechnic Institute (2001).
[30] Soni, Jimmy, and Rob Goodman. A mind at play: how Claude Shannon invented the information age. Simon and Schuster, 2017. p.159
[31] Aguirre, Anthony, Brendan Foster, and Zeeya Merali. "It from bit or bit from it." On Physics And Information. Springer, Berlin (2015).
[32] Losee, Robert M. Information from processes: About the nature of information creation, use, and representation. Springer Science & Business Media, 2012
[33] Janich, Peter. What is information? Vol. 55. U of Minnesota Press, 2018
[34] Rescher, Nicholas. Epistemology: An introduction to the theory of knowledge. SUNY Press, 2012., p.15
[35] Sternberg, Robert J., James C. Kaufman, and Elena L. Grigorenko. Applied intelligence. Cambridge University Press, 2008. p.3
[36] David Robson, The Intelligence Trap: Why Smart People Do Stupid Things and How to Make Wiser Decisions, 2020, p.7
[37] Kanazawa, Satoshi. The intelligence paradox: Why the intelligent choice isn't always the smart one. Vol. 1. New York: Wiley, 2012. p.177
[38] Schwartz, Barry. "The paradox of choice: Why more is less." New York: Ecco, 2004.
[39] Surowiecki, James. The wisdom of crowds. Anchor, 2005
[40] Page, Scott E. The model thinker: what you need to know to make data work for you. Hachette UK, 2018
[41] Nilsson, Nils J. Understanding beliefs. MIT Press, 2014. p.7
[42] Dempster, Arthur P. "The Dempster–Shafer calculus for statisticians." International Journal of approximate reasoning 48.2 (2008): 365-377.
[43] Reineking, Thomas. Belief functions: theory and algorithms. Diss. Universität Bremen, 2014.
[44] Aguirre, Anthony, Brendan Foster, and Zeeya Merali. "It from bit or bit from it." On physics and information (2015).
[45] Wissner-Gross, Alexander D., and Cameron E. Freer. "Causal entropic forces." Physical review letters 110, no. 16 (2013): 168702